\documentclass[journal,10pt,doublecolumn,singlespace]{IEEEtran}
\usepackage{cite,graphicx,subfig,multirow,array}

\usepackage{amsfonts,amssymb}
\usepackage[cmex10]{amsmath}
\usepackage{algorithmic}
\usepackage[ruled]{algorithm2e}	
\usepackage{epstopdf}
\usepackage{hyperref}
\usepackage{verbatim}

\usepackage[usenames,dvipsnames]{color}
\usepackage{soul}

\usepackage{colortbl}
\definecolor{mygray}{gray}{.9}

\begin{document} 
	
\pagenumbering{gobble}

\title{Light Field Denoising via Anisotropic Parallax Analysis in a CNN Framework}

\author{\IEEEauthorblockN{Jie~Chen, Junhui~Hou, and Lap-Pui~Chau}
	
\thanks{J. Chen, and L.-P. Chau are with the School of Electrical \& Electronic Engineering, Nanyang Technological University, Singapore (e-mail: \{Chen.Jie, ELPChau\}@ntu.edu.sg), J. Hou is with the Department of Computer Science, City University of Hong Kong (e-mail: jh.hou@cityu.edu.hk).}
\thanks{The research was partially supported by the ST Engineering-NTU Corporate Lab through the NRF corporate lab@university scheme.}
\thanks{J. Hou was supported in part by the Hong Kong RGC Early Career Scheme Funds 9048123 (CityU 21211518)}
}

\markboth{}
{\MakeLowercase{\textit{et al.}}: }
	
\maketitle

\begin{abstract}
Light field (LF) cameras provide perspective information of scenes by taking directional measurements of the focusing light rays. The raw outputs are usually dark with additive camera noise, which impedes subsequent processing and applications.
We propose a novel LF denoising framework based on anisotropic parallax analysis (APA).
Two convolutional neural networks are jointly designed for the task: first, the structural parallax synthesis network predicts the parallax details for the entire LF based on a set of anisotropic parallax features. These novel features can efficiently capture the high frequency perspective components of a LF from noisy observations. 
Second, the view-dependent detail compensation network restores non-Lambertian variation to each LF view by involving view-specific spatial energies. 
Extensive experiments show that the proposed APA LF denoiser provides a much better denoising performance than state-of-the-art methods in terms of visual quality and in preservation of parallax details.
\end{abstract}


\section{Introduction} \label{sec_intro}

\begin{figure}[t]
	\centering
	\includegraphics[width=0.95\linewidth]{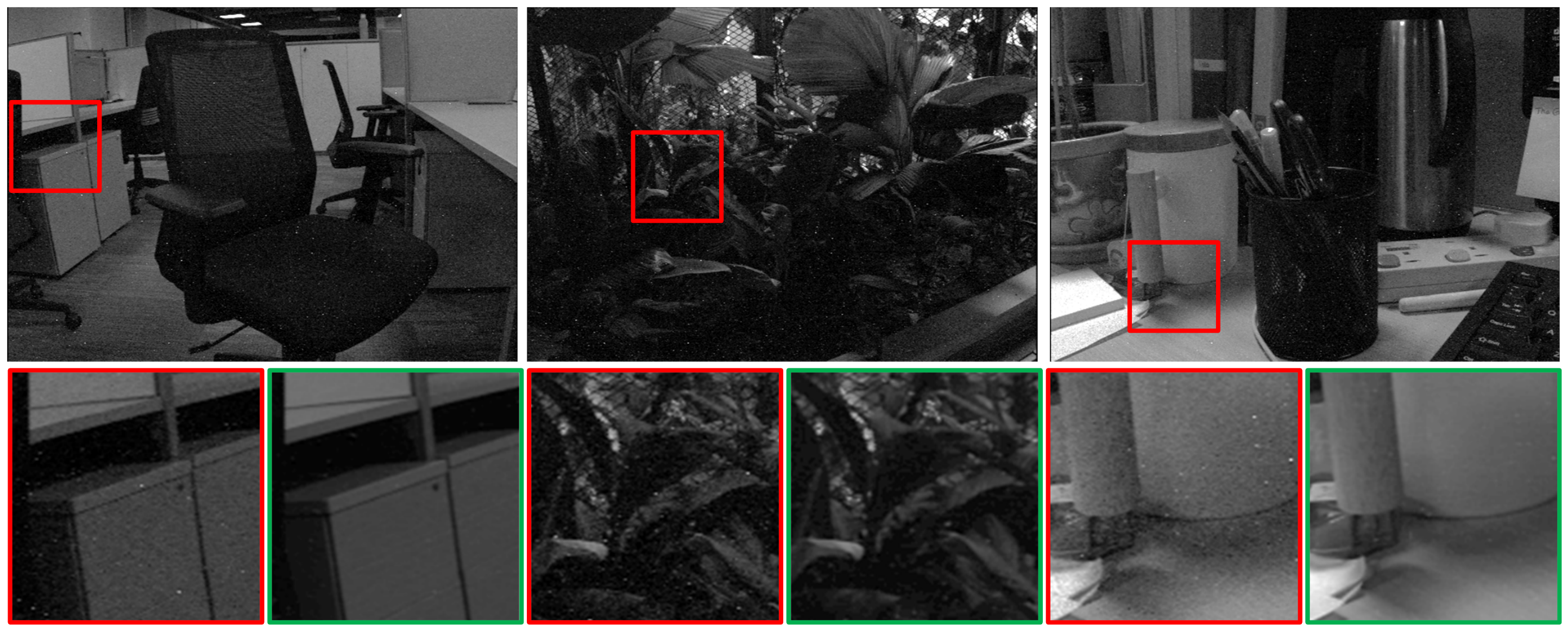}
	\caption{Noisy captures by Lytro Illum. Zoom-in regions in green boxes are the denoising outputs with the proposed \textit{APA} denoiser (estimated noise at $\sigma=10$).}
	\label{fig_realNoise}\vspace{-0.5cm}
\end{figure}

Imaging under low light conditions is challenging as sensor noise can dominate measurements. For commercial grade light field (LF) cameras based on micro-lens arrays \cite{Ng2005} \cite{perwass2012single}, such challenge is particularly pressing. The camera sensor pixel sizes are designed to be small, and the sampling is quite sparse under the micro-lenses \cite{li2013joint}.
Consequently, substantially weaker energy reaches the sensors, making its raw outputs rather noisy even when the scene is well-lit.
Fig. \ref{fig_realNoise} features some noisy captures by Lytro Illum \cite{Ng2005}, where camera noises obviously degrade the image quality. 
Denoising therefore becomes an important procedure for subsequent LF applications \cite{chen2018accurate, shin2018epinet, wang20164d, hou2018light}.

\begin{figure}[t]
	\centering
	\includegraphics[width=0.72\linewidth]{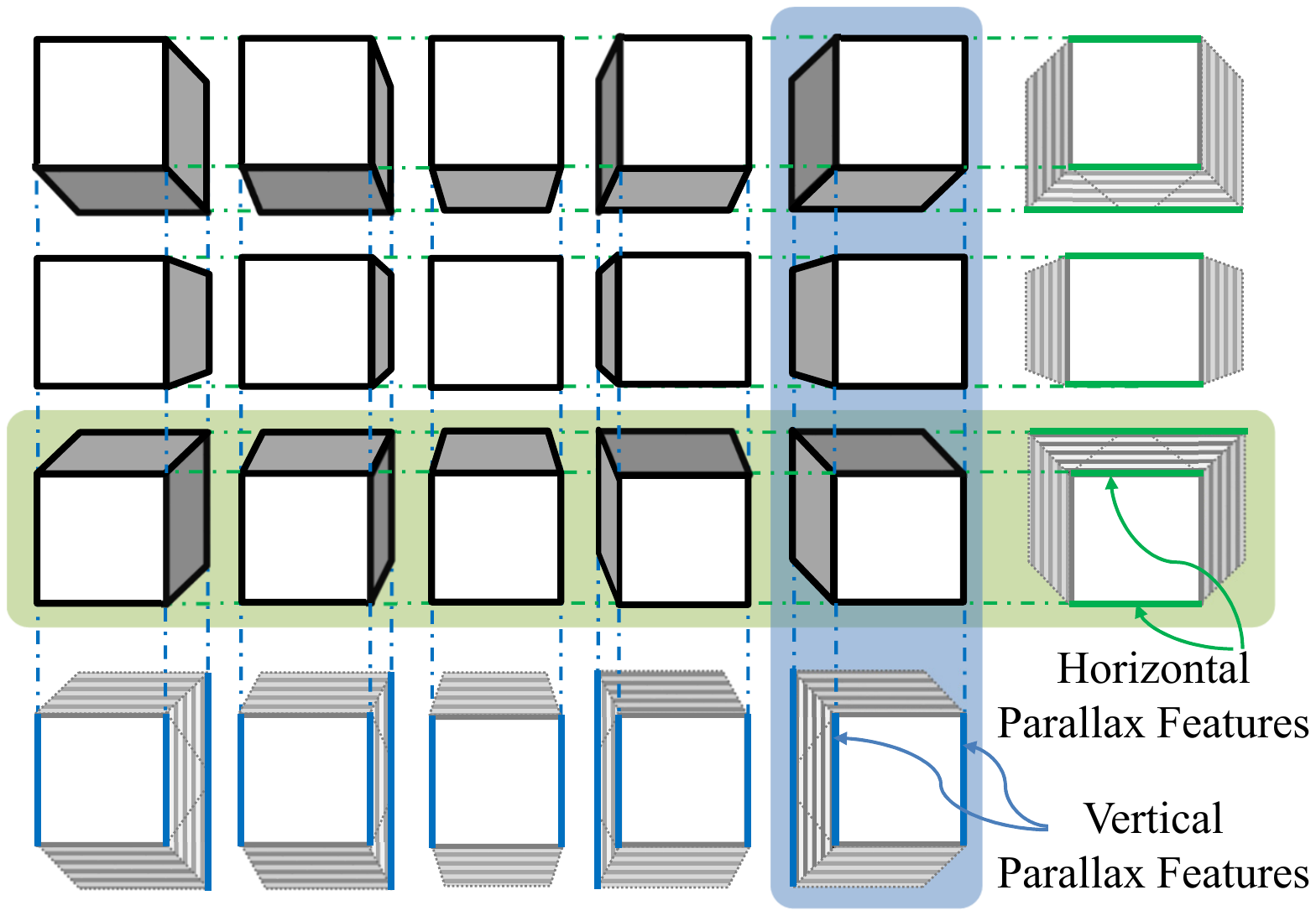}
	\caption{Demonstration of light field parallax variations. The parallax averaging blur is shaded in gray, the vertical and horizontal \textit{anisotropic parallax details} are highlighted in blue and green lines, respectively.}
	\label{fig_parallaxMotion}\vspace{-0.5cm}
\end{figure}

\begin{figure*}[!t]
	\centerline{\subfloat{\includegraphics[width=0.92\linewidth]{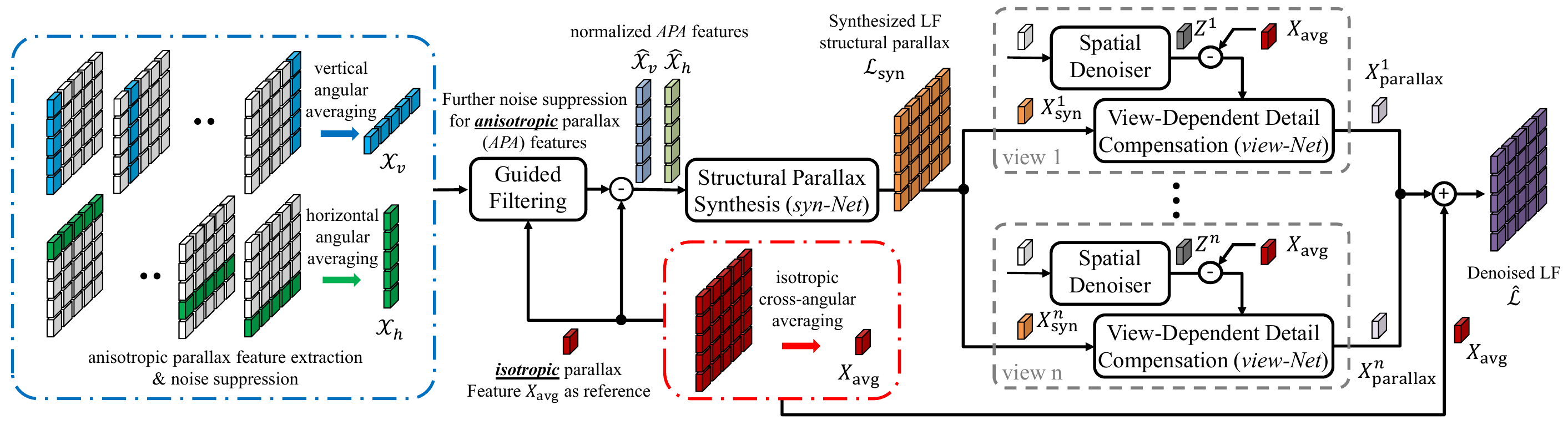}}}
	\caption{System diagram for the proposed \textit{APA} LF denoiser.}
	\label{fig_systemDiagram}\vspace{-0.5cm}
\end{figure*}

One naive approach to light field denoising is to independently employ image denoisers on LF sub-aperture images (SAI) \cite{Ng2005}.
The topic of image denoising has enjoyed extensive development over the last decade \cite{shao2014from}. Statistical \cite{portilla2003image} \cite{sendur2002bivariate} and learning-based methods \cite{Elad2006} \cite{chen2015multiscale} have shown efficiency in the modelling and identification of visual signals from noise. Notably, the classic denoiser \textit{BM3D} \cite{dabov2007image}, which suppresses the noise by seeking signal redundancy among groups of non-locally matched 2D image examples, remains one of the best image denoisers.
However, such a parallel denoising approach is obviously sub-optimal, since the strong correlations embedded in the angular domain of the LF are left unexplored.

By rearranging the LF SAIs in sequential order, video denoisers can be employed to explore the LF angular correlations in the form of inter-frame motion. For instance, the video denoiser \textit{V-BM4D} \cite{maggioni2012video} is suitable for such task, which constructs 3D spatiotemporal volumes with blocks along motion vectors. By enforcing sparsity in the 4th dimension among a group of 3D volumes that show mutually similar patterns, noise can be effectively rejected.
When applied along the EPI dimension instead of angular ones, \textit{V-BM4D} achieves even better desnoising results as demonstrated in \cite{sepas2016light}.

The content parallaxes among different SAIs of an LF are strictly linear, which is much more regularized as compared with the random motion in a video \cite{chen2018light}. 
In the LF context, \textit{non-local} references can be more reliably replaced by angular \textit{local} neighbors.
For Instance, Li et al. \cite{li2013joint} directly employed image denoisers in the LF angular domain, which shows regular linear patterns and can be more efficiently modelled compared with those in spatial domain. 
The angular disparity flow also proves to be useful in the modelling of the LF \cite{sun2016sparse}.

The \textit{HyperFan4D} \cite{dansereau2013decoding} studies the properties of  the LF frequency domain, and uses the LF spatial/angular relationships to restrict the frequency domain to a hyperfan shaped region. However, the exact shape of the hyperfan is highly sensitive to scene content variation, and the noise energy inside the hyperfan cannot be suppressed. 
Alain et al. \cite{Alain2017} proposed a LF denoiser \textit{LFBM5D}, which stacks disparity compensated 4D patches within an local angular window, and seeks sparsity in the 5th dimension. The high dimensional operations by \textit{LFBM5D} prove to be sometimes disadvantageous as compared with \textit{V-BM4D}. The signal redundancy is difficult to be captured, especially when noise level is high.

The Convolutional Neural Networks (CNN) have been proven to be efficient in both high-level vision tasks \cite{krizhevsky2012imagenet} \cite{redmon2016you} as well as in low-level vision applications for capturing signal characteristics \cite{Kim2016accurate,zhang2017beyong}. Image denoising with a CNN was first investigated in \cite{burger2012image}. With a large training dataset, CNN proves to be able to compete with \textit{BM3D}. CNN also proves its efficiency in capturing the spatial-angular structures of the LF in applications such as LF super-resolution \cite{Yoon2017light} and view synthesis \cite{Kalantari2016learning,wu2017light}. 
In this work, we aim at designing an LF denoiser utilizing the CNN's capacities in capturing LF parallax details from noisy observations. A group of anisotropic parallax features are proposed to locally explore the angular domain redundancy under the noise. The structural parallax details as well as view-dependent, non-Lambertian content variations will be restored by the CNN, which is trained over a synthetic noisy LF dataset.

\section{Proposed Algorithm}\label{sec_algo}

Suppose we have a noisy LF observation $\mathcal{L}= \mathcal{\hat{L}}+\mathcal{N}$, where $\mathcal{L}$, $\mathcal{\hat{L}}$, and $\mathcal{N} \in\mathbb{R}^{w\times h\times n_h\times n_v}$ are the noisy LF, noise-free LF, and additive white Gaussian noise (AWGN), respectively. $w$ and $h$ specify the spatial resolutions of each SAI. $n_h$ and $n_v$ specify the horizontal and vertical angular dimensions. We aim to restore $\mathcal{\hat{L}}$ based on $\mathcal{L}$, which consists of noisy SAIs $\mathcal{L}=\{X^n~|~n=1,2,..,n_h\times n_v\}$. For each $X^n$, superscript $n$ indicates the SAI angular index. For simplicity, we denote $X^n$ as $X$ when no confusion exists between different SAIs.

We propose a LF denoising framework via anisotropic parallax analysis (\textit{APA}). The system diagram is shown in Fig. \ref{fig_systemDiagram}, which consists of two sequential CNN modules, i.e., the structural parallax synthesis net (\textit{syn-Net}), and the view-dependent detail compensation net (\textit{view-Net}). 

\subsection{Noise Suppressed Anisotropic Parallax Features for LF Structural Parallax Synthesis}

The LF SAIs represent virtual views that look at a target scene with different perspectives. The viewing locations and directions of the SAIs are configured in a strictly equidistant lattice \cite{Ng2005} \cite{chen2015light}, which results in an array of 2D images that show  parallaxes in arithmetic progression along each angular dimension as illustrated in Fig. \ref{fig_parallaxMotion}.
In such a configuration, the parallaxes show regular directional patterns. For instance, the SAIs along the vertical direction (Highlighted in blue in Fig. \ref{fig_parallaxMotion}) exhibit gradual vertical parallax variations, while no horizontal parallax can be observed among them. When angular averaging is applied among these SAIs, the vertical parallaxes will blur out horizontal high frequency details, while the vertical ones will be preserved as highlighted in the last row of Fig. \ref{fig_parallaxMotion}. Similar blurring processes happen when averaging is carried out among SAIs in horizontal directions.

Let $\Gamma(\mathcal{L},m)$ be the operator to calculate the mean along the $m^\text{th}$ dimension of a tensor $\mathcal{L}$. We calculate the horizontal and vertical \textit{anisotropic parallax} (\textit{APA}) features as:
\begin{align}\label{eq_APA}
\mathcal{X}_h &= \Gamma(\mathcal{L},4),~\mathcal{X}_h\in\mathbb{R}^{w\times h\times n_h}, \notag \\ 
\mathcal{X}_v &= \Gamma(\mathcal{L},3),~\mathcal{X}_v\in\mathbb{R}^{w\times h\times n_v}.\vspace{-0.2cm}
\end{align}
\textit{APA} features $\mathcal{X}_h$ and $\mathcal{X}_v$ extract content details in different directions from the noisy observation $\mathcal{L}$. However, since only a small fraction of SAIs (one row or one column) are used to calculate $\mathcal{X}_h$ and $\mathcal{X}_v$, noise might not be efficiently suppressed, which will deteriorate the accuracy of the extracted features.

To better suppress the noise, we fully utilize all SAIs of the LF and calculate the \textit{isotropic parallax} feature as:
\begin{equation}\label{eq_isotropic}
X_\text{avg}= \Gamma(\Gamma(\mathcal{L},3),4),~X_\text{avg}\in\mathbb{R}^{w\times h}.\vspace{-0.1cm}
\end{equation}
Compared with the \textit{APA} features, $X_\text{avg}$ shows much better noise reduction, however high frequency details are lost in an isotropic manner.
A subsequent guided filtering step \cite{he2010guided} is introduced to transfer the high frequency parallax details of $\mathcal{X}_h$ and $\mathcal{X}_v$ onto the isotropic reference $X_\text{avg}$. 
Let $\mathcal{G}(I,J)$ denote the guided filtering over $J$ using $I$ as guide, and the \textit{normalized APA features} are calculated as:
\begin{align}
\mathcal{\hat{X}}_h(:,:,n)= \mathcal{G}(\mathcal{X}_h(:,:,n), X_\text{avg})- X_\text{avg},\notag\\
\mathcal{\hat{X}}_v(:,:,n)= \mathcal{G}(\mathcal{X}_v(:,:,n), X_\text{avg})- X_\text{avg}.
\end{align}

Based on these normalized and ideally \textit{noise-free APA} features $\{\mathcal{\hat{X}}_h,\mathcal{\hat{X}}_v\}\in\mathbb{R}^{w\times h\times (n_h+n_v)}$, a CNN (denote as \textit{syn-Net}) is used to reconstruct the entire LF parallax structure $\mathcal{L}_\text{syn}\in\mathbb{R}^{w\times h\times n_h\times n_v}$. $\mathcal{L}_\text{syn}$ contains the parallax details for each SAI and restores their spatial perspective relationships.

\subsection{View-Dependent Detail Compensation}

\begin{figure}[t]
	\centering
	\includegraphics[width=0.96\linewidth]{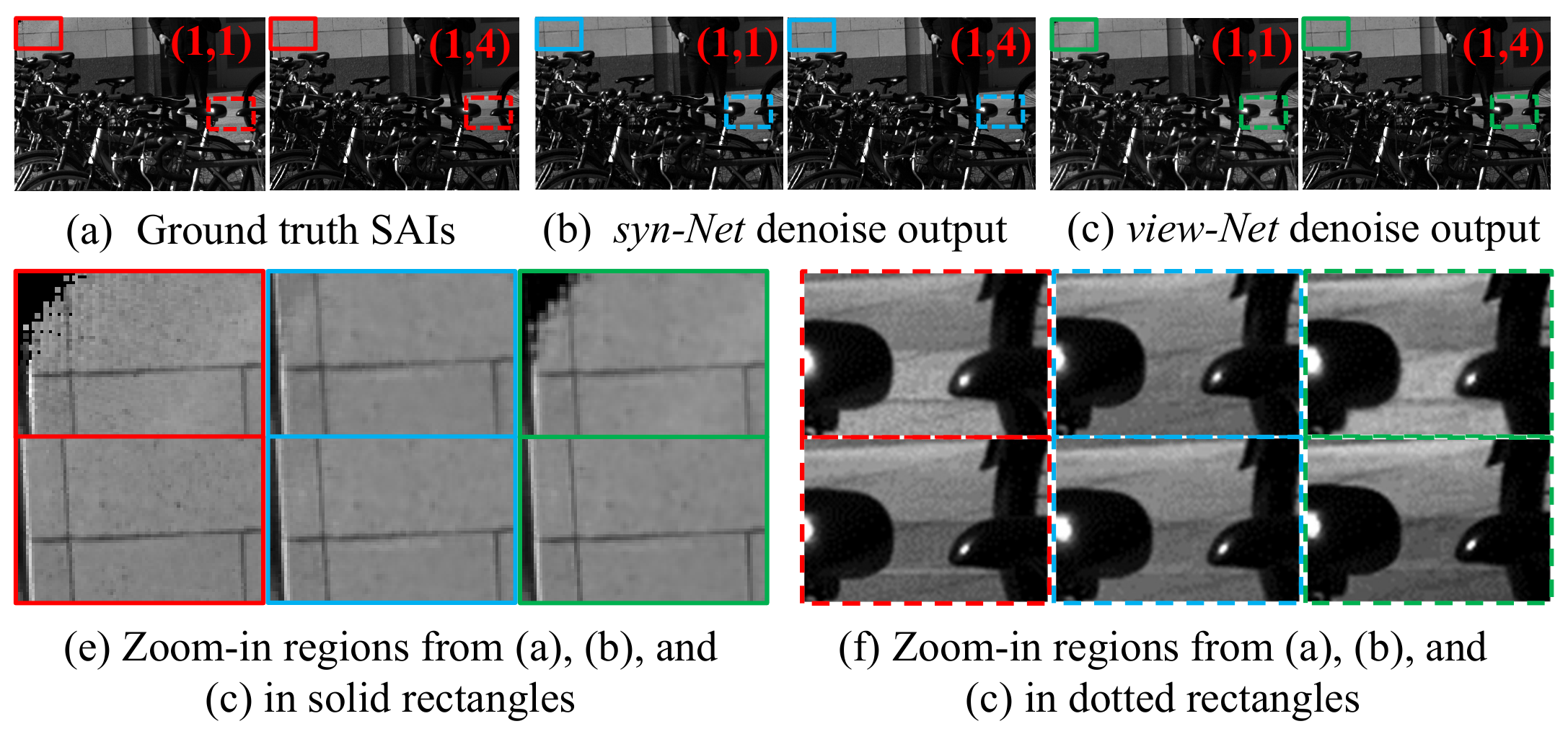}
	\caption{Demonstration of view-dependent details restored by \textit{syn-Net} and \textit{view-Net}. The (1,1)-st and (1,4)-th SAI from a 8$\times$8 LF are used for comparison.}
	\label{fig_visualComp_viewDependent}\vspace{-0.5cm}
\end{figure}

The \textit{syn-Net} predicts the structural parallax details of the entire LF in one forward pass. Its convolutional filters (except last layer) are shared among all SAIs, and non-Lambertion variations among the SAIs will be neutralized during training. Consequently, the output SAIs $X_\text{syn}^n$ ($X_\text{syn}^n\in\mathcal{L}_\text{syn},n=1,2,...$) are always Lambertian. As illustrated in Fig. \ref{fig_visualComp_viewDependent}(e) and (f), view-dependent, non-parallax differences between the SAIs caused by factors such as directional lighting, lens vignetting, and non-Lambertian/specular reflective surfaces, cannot be efficiently restored by the \textit{syn-Net}. 

Being an informative part of an LF, view-dependent uniqueness should be truthfully preserved.
To incorporate view-specific details to the network, we use a 2D Gaussian spatial filter to suppress the noise of each SAI independently, and the denoised SAI is denoted as $Z^n\in\mathbb{R}^{w\times h}$ ($n=1,2,..,n_h\times n_v$). 
The low-pass Gaussian filter partially removes the noise and the high frequency content details, however preserves low frequency view-dependent local energies. 

A subsequent module of parallel CNNs, which we denote as \textit{view-Net}, will be used for every SAI based on the input features $\{(Z^n\text{-}X_\text{avg}),X_\text{syn}^n\}\in\mathbb{R}^{w\times h\times 2}$. The \textit{view-Net} output $X_\text{parallax}^n\in\mathbb{R}^{w\times h}$ is supposed to have embedded both structural parallax details and view-dependent local energies. As shown in Fig. \ref{fig_systemDiagram}, all SAIs are processed with identical CNN architectures of the \textit{view-Net} to form the final denoised LF $\mathcal{\hat{L}}$.

\subsection{Network Specifics and Training Details}

\begin{figure}[t]
	\centering
	\includegraphics[width=0.9\linewidth]{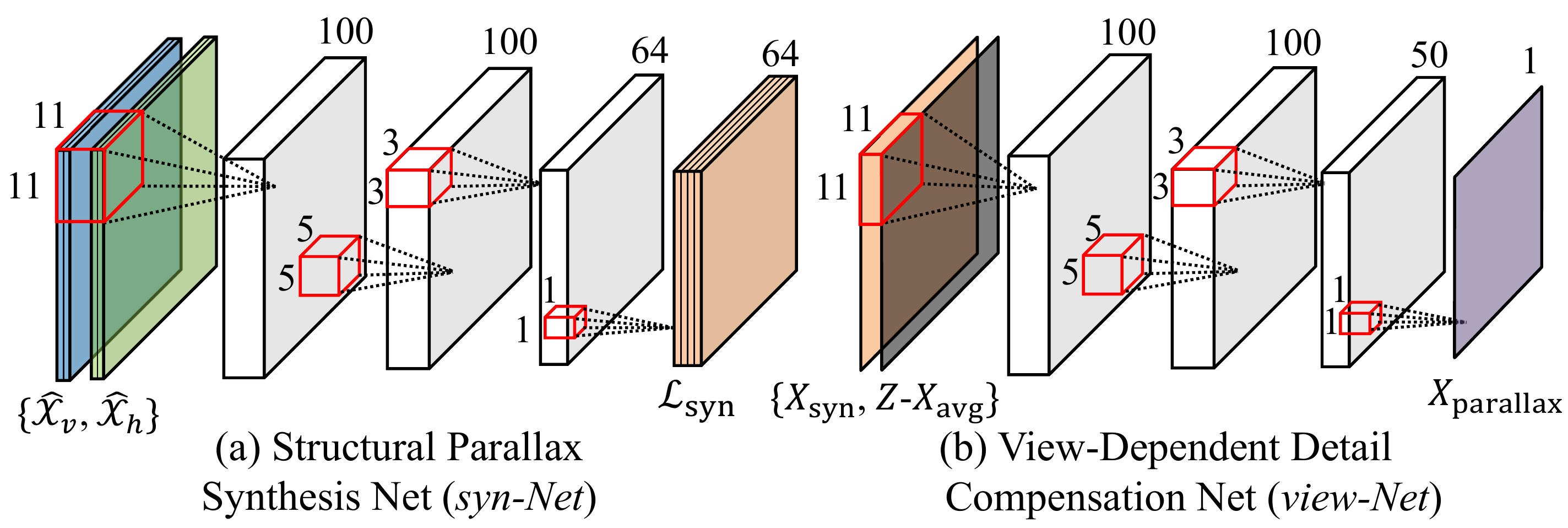}
	\caption{CNN structures of the proposed \textit{APA} LF denoiser. Each convolutional layer is followed by a ReLU layer.}
	\label{fig_CNNStructure}\vspace{-0.2cm}
\end{figure}

The CNN architectures of the \textit{syn-Net} and the \textit{view-Net} are shown in Fig. \ref{fig_CNNStructure}. Both nets consist of four convolutional layers with decreasing kernel sizes of 11, 5, 3, and 1. Each layer is followed by a rectified linear unit (ReLU). 

The two CNNs were trained separately. First, the \textit{syn-Net} was trained based on the cost function:\vspace{-0.2cm}
\begin{equation}
E_\text{syn}=[\mathcal{\hat{L}}- \mathcal{L}_{syn}]^2,
\end{equation}
where $\mathcal{\hat{L}}$ denotes the ground truth LF. After the \textit{syn-Net} had converged, the \textit{view-Net} was trained for each SAI independently based on another cost function:
\begin{equation}\vspace{-0.2cm}
E_\text{view}=[\hat{X}- X_\text{parallax}- X_\text{avg}]^2.
\end{equation}
Here $\hat{X}$ is one SAI of the ground truth LF, and the \textit{view-Net} output $X_\text{parallax}$ is calculated with $X_\text{syn}$ as input, which is one SAI of the \textit{syn-Net} output $\mathcal{L}_\text{syn}$. Stochastic gradient descent (SGD) was used to minimize the objective function. Mini-batch size was set as 50. The Xavier approach \cite{glorot2010understanding} was used for network initialization, and the ADAM solver \cite{kingma2014adam} was adopted for system training, with parameter settings $\beta_1=$ 0.9, $\beta_2=$ 0.999, and learning rate $\alpha=$ 0.0001.

\begin{table}[t]
	\begin{center}
		\small\centering\setlength\tabcolsep{1pt}
		\caption{LF Denoising Performance comparison in terms of PSNR and SSIM among different methods.}
		\vspace{-0.2cm} 
		\label{tbl_psnrSSIM}
		\begin{tabular}{|>{\centering\arraybackslash}m{0.6cm}|>{\centering\arraybackslash}m{2.4cm}|>{\centering\arraybackslash}m{1.7cm}|>{\centering\arraybackslash}m{1.7cm}|>{\centering\arraybackslash}m{1.7cm}|}
			\hline 
			\multicolumn{2}{|c|}{\multirow{3}{*}{Method}} &\multicolumn{3}{c|}{\textit{noise level} $\sigma\in[0,255]$} \\ \cline{3-5} 
			\multicolumn{2}{|c|}{} &10 &20 &50 \\ \cline{3-5}
			\multicolumn{2}{|c|}{} &PSNR/SSIM &PSNR/SSIM &PSNR/SSIM \\ \hline\hline				
			\multicolumn{2}{|c|}{\textit{BM3D} \cite{dabov2007image}} &36.53/ 0.93 &33.04/ 0.87 &28.75/ 0.75\\ \hline			
			\multicolumn{2}{|c|}{\textit{HyperFan4D} \cite{dansereau2013light}} &32.06/ 0.89 &30.23/ 0.79 &27.94/ 0.68 \\ \hline								
			\multicolumn{2}{|c|}{\textit{V-BM4D} \cite{maggioni2012video}} &38.33/ 0.95 &34.89/ 0.90 &30.53/ 0.79\\ \hline
			\multicolumn{2}{|c|}{\textit{V-BM4D-EPI} \cite{sepas2016light}} &38.45/ 0.96 &35.32/ 0.92 &31.24/ 0.82\\ \hline
			\multicolumn{2}{|c|}{\textit{LFBM5D} \cite{Alain2017}} & 36.98/ 0.92 & 33.25/ 0.85 & 28.10/ 0.73 \\ \hline\hline
			\multirow{3}{*}{\rotatebox[origin=c]{90}{Ours}} &\textit{Avg-All} &30.57/ 0.88 &29.91/ 0.85 &27.61/ 0.67\\\cline{2-5}
			&\textit{APA-syn}&36.43/ 0.96 &35.49/ 0.94 &33.41/ 0.90\\ \cline{2-5}
			&\textit{APA}&\textbf{38.80}/ \textbf{0.97} &\textbf{36.60}/ \textbf{0.95} &\textbf{33.77}/ \textbf{0.91}\\ \hline
		\end{tabular}
	\end{center}
	\vspace{-0.6cm}
\end{table}

\begin{figure*}[!t]
	\centerline{\subfloat{\includegraphics[width=0.96\linewidth]{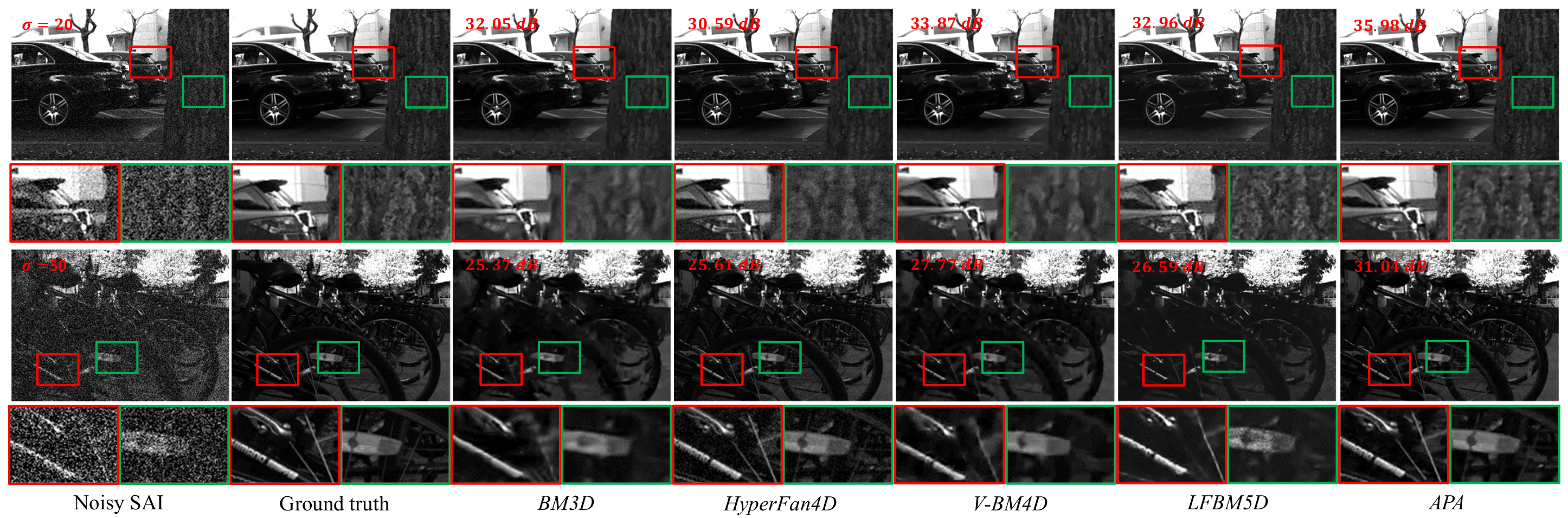}}}
	\caption{Visual comparison of denoised LFs from different methods. The $(4,4)$-th SAI is shown for each LF. PSNR for each denoised LF is shown at the top left of each image. Digital zoom-in is suggested for detail comparison.}
	\label{fig_visualComp_denoise}\vspace{-0.4cm}
\end{figure*}

To create the synthetic noisy LF dataset for training, we chose 70 scenes from the Stanford Lytro Light Field Archive \cite{stanfordArchive}. For each LF scene, the central 8$\times$8 (out from 14$\times$14) SAIs were extracted and converted to gray-scale for evaluation (though the \textit{APA} framework can be easily extended to RGB space). Zero-mean AWGN with standard variance of $\sigma= 10$, $20$, and $50$ were synthesized to each training data. Patches of size $32\times 32$ were extracted from each LF with stride of 16. Finally we have around 50,000 patches for training at each noise level. Note that 3 separate networks were trained and evaluated independently at each noise level. In case of an input with arbitrary noise, the network closest to the estimated noise can be used. In real world applications, more networks could be trained at a closer noise interval. The details of datasets and evaluation results can be found online\footnote{Dataset details and complete evaluation results available via: \url{https://github.com/hotndy/APA-LFDenoising}\label{footnote1}}.

\section{Experimental Results} \label{sec_exp}

In this section, we evaluate the performance of the proposed LF denoiser \textit{APA}, and compare with state-of-the-art LF denoisers. 30 LF scenes (different from the training dataset) were randomly picked from the Stanford Lytro Light Field Archive \cite{stanfordArchive}. Zero-mean AWGN of standard variance $\sigma=10$, $20$, and $50$ was synthesized onto these scenes as the testing dataset\footnotemark[\value{footnote}].

Five methods were chosen for comparison: the classic 2D image denoiser \textit{BM3D} \cite{dabov2007image}, which denoises each SAI independently; the video denoiser \textit{V-BM4D} \cite{maggioni2012video}, which considers the LF SAIs as a video sequence, and its variation \textit{V-BM4D-EPI}\cite{sepas2016light} that works on the EPI sequence; the LF denoiser \textit{HyperFan4D} \cite{dansereau2013decoding}, which works on the 4D frequency space of the LF; and \textit{LFBM5D} \cite{Alain2017}, which extends the BM3D model into a 5D framework. Two baseline methods were used for evaluation: \textit{Avg-All}, which angularly replicates $X_\text{avg}$ to reconstruct the entire LF; \textit{APA-syn}, which uses $\mathcal{L}_\text{syn}$ from the \textit{syn-Net} as denoising output. 

\begin{figure}[t]
	\centering
	\includegraphics[width=1\linewidth]{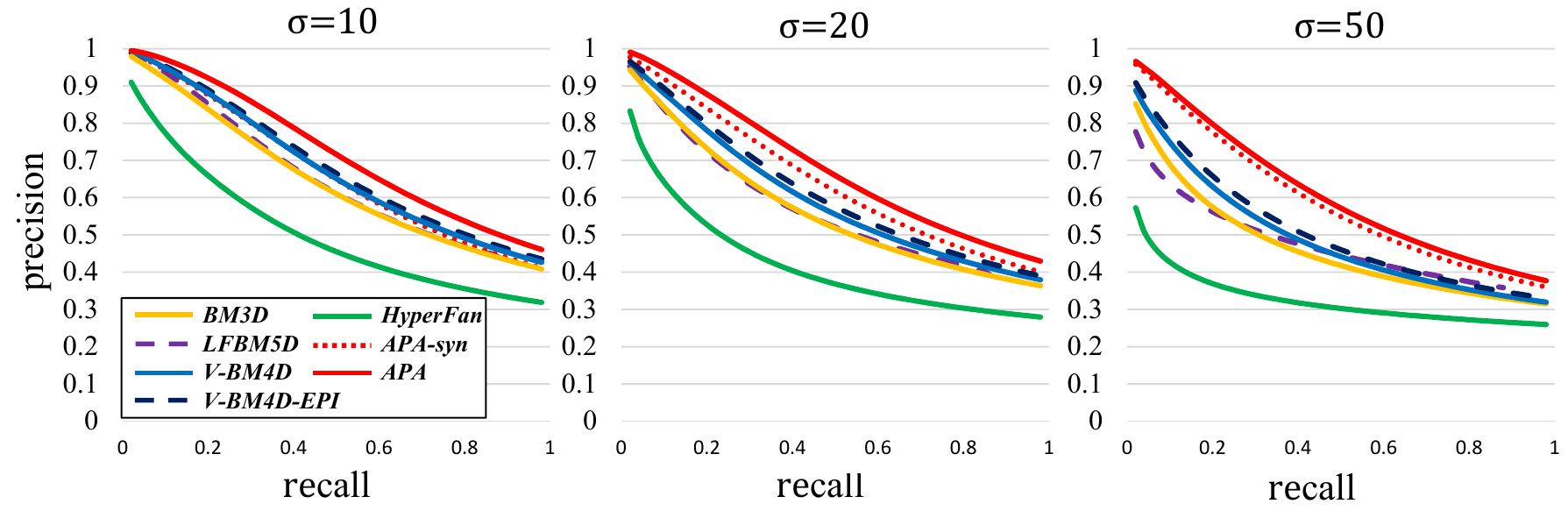}
	\caption{Parallax content precision recall curves for different methods under different noise levels.}
	\label{fig_prCurves}\vspace{-0.5cm}
\end{figure}

\subsection{Model Quantitative Evaluation}

We quantitatively evaluated the proposed \textit{APA} denoiser. All the results reported in this section are average of all LF SAIs.

\subsubsection{\textbf{PSNR/SSIM of Denoised LF}}
We calculated the PSNR and Structural Similarity (SSIM) \cite{wang2004image} between the ground truth LFs and the denoised ones for different methods over the testing dataset, and the average results for all testing data are shown in TABLE \ref{tbl_psnrSSIM}. As can be seen, \textit{APA} shows better denoising performance than all the competing methods at all noise levels. The advantage is especially obvious under larger noise levels: at $\sigma=50$, 5.8 \textit{dB} and 3.2 \textit{dB} advantages are achieved by \textit{APA} over \textit{HyperFan4D} and \textit{V-BM4D}, respectively. 

The baseline \textit{syn-Net} output \textit{APA-syn} already shows better performance than all other competing denoisers at $\sigma=20$ and $50$. The marginal advantage from the subsequent \textit{view-Net} is largest at $\sigma=10$ (2.4 \textit{dB}), and smaller at $\sigma=50$ (0.4 \textit{dB}).
 
\subsubsection{\textbf{LF Parallax Preservation}}

The most important information embedded in an LF image are the parallaxes among different SAIs. How well this information is preserved/restored by the denoiser should be an important evaluation metric.
To this end, we propose to calculate the parallax content precision-recall (PR) curves. First, the central SAI is subtracted from all SAIs of an LF. Different threshold values are then applied to produce a binary \textit{parallax edge map} for each SAI. We compared the binary parallax edge maps between the ground truth LF and the denoised LF to plot the PR curves for each method, and the corresponding results are shown in Fig. \ref{fig_prCurves}. The PR curves give an intuitive assessment on how well the parallax is preserved. As can be seen, the proposed \textit{APA} denoiser can best restore the LF parallax. 

\subsection{Visual Quality Evaluation}

Fig. \ref{fig_visualComp_denoise} gives visual comparisons of the denoised LF. The $(4,4)$-th SAI of the denoised LFs are shown for each method. As can be seen, the \textit{APA} denoiser proves to best remove the noise and restore the scene details. Both \textit{BM3D} and \textit{V-BM4D} seriously blurs the image under large noise levels. Noticeable remaining noise can be found in the output of \textit{HyperFan4D} and \textit{LFBM5D}. Fig. \ref{fig_realNoise} shows the denoising results over direct noisy inputs from Lytro Illum (estimated noise range $\sigma=10$). As can be seen, although the \textit{APA} denoiser has been trained over Gaussian noises, it generalizes well to real camera noises.

\section{Concluding Remarks} \label{sec_conclusion}

We have proposed a novel LF denoising framework based on anisotropic parallax analysis. Based on the \textit{APA} features, two sequential CNNs have been designed to first create the structural parallax details, and then restore view-dependent local energies. 
Extensive experiments show that the proposed \textit{APA} LF denoiser provides a better denoising performance than state-of-the-art methods in terms of visual quality and in preservation of parallax details between the views. 

The calculation of noise-free \textit{APA} features involves direct angular averaging of the SAIs. This step is vulnerable to noise of large magnitudes. Additionally, the spatial denoiser in the \textit{view-Net} is a simple Gaussian low-pass filter, which can create obvious artifacts. These facts on the one hand further validates the capacity of the CNNs in dealing with imperfect inputs; on the other hand, it shows great improvement potential of the proposed \textit{APA} LF denoiser. We will further investigate these potentials in the future.


\bibliographystyle{IEEEbib}
\bibliography{refs}

\end{document}